\documentclass[11pt]{article}

\usepackage[preprint]{acl}

\usepackage{times}
\usepackage{latexsym}

\usepackage[T1]{fontenc}

\usepackage[utf8]{inputenc}

\usepackage{microtype}

\usepackage{inconsolata}

\usepackage{graphicx}

\usepackage{amsmath}
\usepackage{amssymb}

\title{Modelling the Diachronic Emergence of Phoneme Frequency Distributions}

\author{Fermín Moscoso del Prado Martín \\
  Department of Computer Science\\ and Technology \\
  University of Cambridge, UK \\
  \texttt{fm611@cst.cam.ac.uk} \\
  \And
  Suchir Salhan \\
  Department of Computer Science\\ and Technology \\
  University of Cambridge, UK \\
  \texttt{sas245@cst.cam.ac.uk} \\}

\begin{document}
\maketitle
\begin{abstract}
Phoneme frequency distributions exhibit robust statistical regularities across languages, including exponential-tailed rank-frequency patterns and a negative relationship between phonemic inventory size and the relative entropy of the distribution. The origin of these patterns remains largely unexplained. In this paper, we investigate whether they can arise as consequences of the historical processes that shape phonological systems. We introduce a stochastic model of phonological change and simulate the diachronic evolution of phoneme inventories. A naïve version of the model reproduces the general shape of phoneme rank-frequency distributions but fails to capture other empirical properties. Extending the model with two additional assumptions --an effect related to functional load and a stabilising tendency toward a preferred inventory size-- yields simulations that match both the observed distributions and the negative relationship between inventory size and relative entropy. These results suggest that some statistical regularities of phonological systems may arise as natural consequences of diachronic sound change rather than from explicit optimisation or compensatory mechanisms.
\end{abstract}

\section{Introduction}

The frequency distributions with which different phonemes occur in a language can provide insights into the nature, representation, and processing of human language. Nevertheless, despite its importance, only a handful of studies have investigated the nature of these distributions; most investigated these distributions from a macroscopic (in the sense of \citealp{Mandelbrot:1957}) perspective, considering the shape of the rank-frequency plots \citep{Sigurd:1968,Good:1969,Martindale:etal:1996,Martindale:Tambovtsev:2007,MacklinCordes:Round:2020,Moscoso:Salhan:2026}. 

Recently, it has been reported \citep{Moscoso:Salhan:2026} that --at both macro- and microscopic levels of description-- phoneme frequency distributions exhibit detectable effects consistent with the `Compensation Hypothesis' \citep{Hockett:1955,Martinet:1955}. This hypothesis predicts that increased complexity in one domain of language is offset elsewhere in the language. Several studies had previously found evidence for this hypothesis involving balancing aspects of the phonological system with other aspects of language structure \citep{Moran:Blasi:2014,Pimentel:etal:2020,Pimentel:etal:2021}. It is, however, remarkable that compensation is already detectable by examining the unigram phoneme frequency distributions alone.

An important related question has received little attention: How do phoneme frequency distributions arise in diachronic terms? In particular, one could also ask whether what appear as compensation effects might in fact be unexpected consequences of the historical processes that have shaped the phoneme inventories. This would open the possibility of compensation phenomena being epiphenomenal, rather than the result of any actual optimisation process. \citet{Ceolin:etal:2019} and \citet{Ceolin:2020} introduce a stochastic split-and-merger model of sound change in which phoneme inventories evolve through repeated splitting and merging events. Their simulations show that statistical patterns commonly associated with phonological markedness can emerge from simple diachronic processes, without assuming markedness as a primitive property of phonological systems. In a similar vein, in this study, we introduce a model of phoneme change based on \citeposs{Hoenigswald:1965} typology of phonological changes. We use this model to investigate whether two simple diachronic principles --functional load and a stabilising tendency in phoneme inventory size-- are sufficient for generating both the observed phoneme rank-frequency distributions and the negative relationship between phonemic inventory size and relative entropy. 

In what follows, we first describe the empirical properties of phoneme frequency distributions that our model aims to explain (Section~2). We then introduce a stochastic model of phonological change (Section~3). We examine three incremental versions of the model of increasing complexity: a naïve baseline (Section~3.3), a version incorporating effects related to functional load (Section~3.4), and a model further introducing a central tendency in phonemic inventory size (Section~3.5). We show that the latter reproduces both the observed rank-frequency patterns of phoneme frequencies and the negative relationship between phonemic inventory size and relative entropy, suggesting that these macroscopic patterns may arise as natural consequences of diachronic phonological dynamics.

\section{Properties of Phoneme Distributions}

\begin{figure*}[t]
\begin{tabular}{ccc}
{\bf (a)} & {\bf (b)} \\
  \includegraphics[width=.48\textwidth]{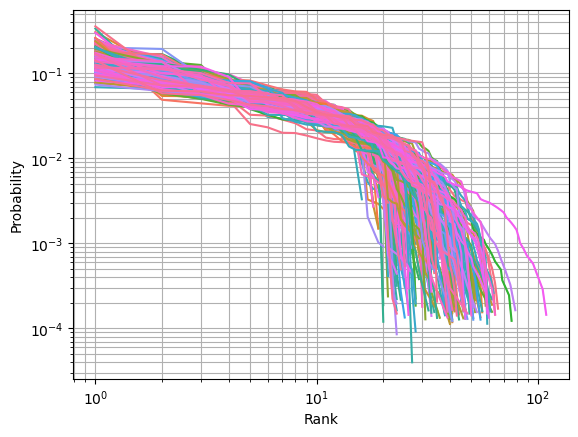} &
  \includegraphics[width=.48\textwidth]{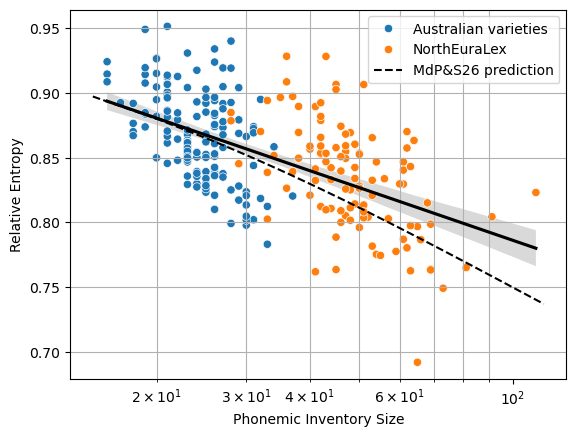} 
\end{tabular}
  \caption {{\bf (a)} Rank-frequency plots (note the log-log scale) for the phoneme frequency distributions across the languages in \citeposs{MacklinCordes:Round:2020} dataset of Australian language varieties, and those in NorthEuraLex. Each line plots one language. {\bf (b)} Relationship between PIS and relative entropy for the languages in \citeposs{MacklinCordes:Round:2020} dataset of Australian language varieties, and those in NorthEuraLex. Each point plots one language. The solid line plot a log-linear regression, and the shading plots its 95\% C.I.. The dashed line plots the relation as predicted by \citet{Moscoso:Salhan:2026}.}
   \label{fig:reallang}
\end{figure*}

\paragraph{Exponential tails:}

Contrary to arguments for power-law distributions of phonemes \citep{Sigurd:1968,Martindale:etal:1996,Martindale:Tambovtsev:2007,Ceolin:etal:2019,Ceolin:2020}, phoneme frequency distributions exhibit rank-frequency plots typical of exponential-tailed distributions \citep{Good:1969,MacklinCordes:Round:2020,Moscoso:Salhan:2026}: In the double-logarithmic plane, the right tails of these distributions fall abruptly, quickly deviating from the straight lines that are characteristic of power-law-tailed distributions (e.g., word frequency distributions; \citealp{Condon:1928}). These patterns are illustrated in Figure~\ref{fig:reallang}a for the 166 languages of \citeposs{MacklinCordes:Round:2020} dataset of Australian language varieties,\footnote{Downloaded from \url{https://zenodo.org/records/4104116} on May 1, 2025.} and for the 107 languages in the NorthEuraLex database \citep{Dellert:etal:2019}.\footnote{v0.9, downloaded from the LexiBank Project at \url{https://github.com/lexibank/northeuralex} on March 3, 2026.}

\paragraph{Correlation between phonemic inventory size and relative entropy:}

In a recent study, \citet{Moscoso:Salhan:2026} report that --across the world's languages-- there is a negative correlation between a language's \emph{Phonemic Inventory Size} ({PIS}), and the relative entropy \cite{Shannon:1948} of its phoneme distribution. The entropy of the phoneme distribution is an indicator of the per-phoneme informational content of a language \citep{Cherry:etal:1953}. Generally speaking, increasing the PIS increases the potential value of the distribution's entropy. However, the reduction in relative entropy attenuates the effect of increasing the PIS on the inventory's entropy, as would be predicted by the Compensation Hypothesis \citep{Hockett:1955,Martinet:1955}. This is illustrated in Figure~\ref{fig:reallang}b. The figure plots the negative correlation between PIS and relative entropy across the languages in the Australian languages and NorthEuraLex datasets mentioned above. Note that this relation is rather robust, it is present on the Australian language dataset --which was one of the datasets used by Moscoso del Prado Mart\'{\i}n and Salhan--, however, the relation also extends to the NorthEuraLex languages --not used in the previous study. In fact, as shown in the figure, the precise prediction made by Moscoso del Prado Mart\'{\i}n and Salhan (dashed line in the figure) extrapolates into the NorthEuraLex observations remarkably well.

\section{Modelling Phonological Change}

\subsection{Phonological changes}

\emph{Phonological changes} (also known as phonemic changes) are diachronic changes within a language’s system of contrastive sounds. Such changes go beyond mere shifts in pronunciation, which are referred to as \emph{phonetic changes}. They affect the oppositions among phonemes, and may result in changes to a language's phoneme inventory: Contrasts may arise, disappear, or be reorganised, resulting in a restructuring of the language’s sound system. 

Phonological changes can be of different types. Using \citeposs{Hoenigswald:1965} systematisation, we can distinguish between three main types: \emph{Primary splits} (also known as conditioned mergers) occur when some instances of a phoneme $A$ become an already existing phoneme $B$ in particular contexts. In these cases, the number of phonemes remains the same, but their distribution changes. \emph{Secondary splits} (also known as phonemic splits) occur when some instances of $A$ become a new phoneme $B$. Changes of this type result in new contrast being created. In turn this increases the size of phonemic inventories. Finally, \emph{unconditioned mergers} (also known simply as mergers) occur when all instances of phonemes $A$ and $B$ collapse into a single phoneme $A$. In these cases, a contrast is eliminated and the number of phonemes decreases. Note that, within this classification, there is no specific type for phoneme losses; these are just considered unconditional mergers with a zero phoneme.

\subsection{The general model}
\label{sec:formal}

Consider  language at a particular time point $\tau$ in its history. It uses $V_\tau$ distinct contrastive phonemes $\pi_1,\pi_2,\ldots,\pi_{V_\tau}$. The probabilities with which each phoneme occurs in the language at time $\tau$ are given by the vector $\boldsymbol{\mathrm{p}}_\tau = \{\mathrm{p}_\tau(\pi_1),\mathrm{p}_\tau(\pi_2),\ldots,\mathrm{p}_\tau(\pi_{V_\tau})\}$, where $\sum_{i=1}^{V_\tau}\mathrm{p}_\tau(\pi_i)=1$.  Time is treated as a discrete sequence of phonological changes $c_\tau$. Each change is of one of three types: primary split ($p$), secondary split ($s$), or merger ($m$). Formally, $c_\tau$ is sampled at each time step from the alphabet $\Sigma=\{p,s,m\}$. Time intervals are defined so that each contains exactly one change event.

We construct stochastic models in which, at every time point $\tau$, the change type $c_\tau$ is sampled randomly from $\Sigma$, with probabilities $P_\tau(p)$, $P_\tau(s)$, and $P_\tau(m)$, so that $P_\tau(p)+P_\tau(s)+P_\tau(m)=1$. Independently, a phoneme $\pi_i \in \{\pi_1,\ldots,\pi_{V_\tau}\}$ is sampled (according to some predefined scheme, details below) as the target of change. In the case of primary splits and mergers, a second phoneme $\pi_j$ is also sampled, representing the phoneme toward which $\pi_i$ shifts (primary split) or with which it merges (merger). In addition, a proportion parameter $\alpha_\tau \in (0,1]$ is sampled to determine how much probability mass is transferred between phonemes. We assume $\alpha_\tau$ is uniformly distributed, reflecting the fact that phonological changes can affect contrasts to varying degrees.

The three types of changes are defined as transformations on the probability vector:
\paragraph{Primary split:} a proportion $\alpha_\tau$ of $\mathrm{p}(\pi_i)$ is reassigned to an existing phoneme $\pi_j$:
\begin{equation} \label{eq:primary}
\begin{aligned}
\mathrm{p}_{\tau+1}(\pi_i) &= (1-\alpha_\tau)\mathrm{p}_\tau(\pi_i),\\
\mathrm{p}_{\tau+1}(\pi_j) &= \mathrm{p}_\tau(\pi_j)+\alpha_\tau\,\mathrm{p}_\tau(\pi_i).
\end{aligned}
\end{equation}
For $0<\alpha_\tau<1$, the inventory size remains $V_{\tau+1}=V_\tau$; if $\alpha_\tau=1$, $\pi_i$ disappears and $V_{\tau+1}=V_\tau-1$.

\paragraph{Secondary split:} a new phoneme $\pi_{V_\tau+1}$ is created by splitting $\pi_i$:
\begin{equation} \label{eq:secondary}
\begin{aligned}
\mathrm{p}_{\tau+1}(\pi_i) &= (1-\alpha_\tau)\mathrm{p}_\tau(\pi_i),\\
\mathrm{p}_{\tau+1}(\pi_{V_\tau+1}) &= \alpha_\tau\,\mathrm{p}_\tau(\pi_i).
\end{aligned}
\end{equation}
For $0<\alpha_\tau<1$, $V_{\tau+1}=V_\tau+1$; if $\alpha_\tau=1$, $\pi_i$ disappears and $V_{\tau+1}=V_\tau$

\paragraph{Unconditioned merger:} two phonemes $\pi_i$ and $\pi_j$ collapse completely:
\begin{equation} \label{eq:merge}
\mathrm{p}_{\tau+1}(\pi_j)
=
\mathrm{p}_\tau(\pi_j)
+
\mathrm{p}_\tau(\pi_i),
\end{equation}
and $\pi_i$ is removed, therefore $V_{\tau+1}=V_\tau-1$. 

Over time, this process generates stochastic trajectories both in inventory size $V_\tau$ and in the internal distributional structure of the phonological system (the probability vector). As we show below, the method by which the change probabilities $P_\tau(p)$, $P_\tau(s)$, and $P_\tau(m)$, the choice of phonemes, and the $\alpha_\tau$ are chosen at each time point can substantially affect the behaviour of these models. In summary, this model treats phonological change as a stochastic process redistributing probability mass across phonemes while allowing contrasts to be created or eliminated.

\begin{figure*}[th]
\begin{tabular}{cc}
{\bf (a)} & {\bf (b)} \\
  \includegraphics[width=.48\textwidth]{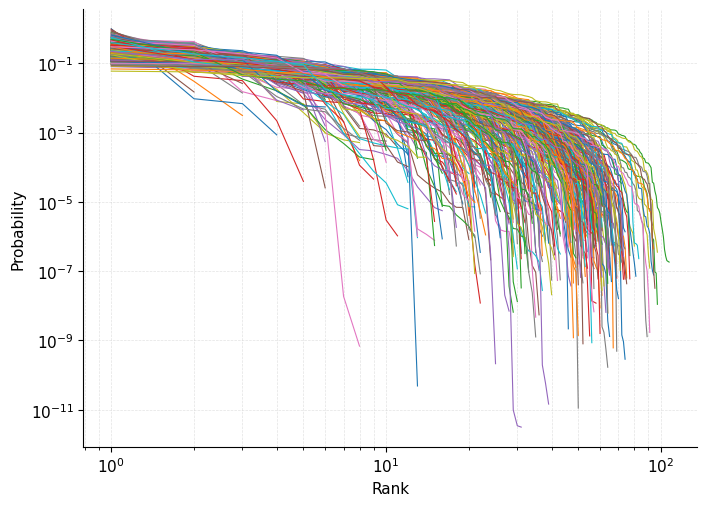} &
  \includegraphics[width=.48\textwidth]{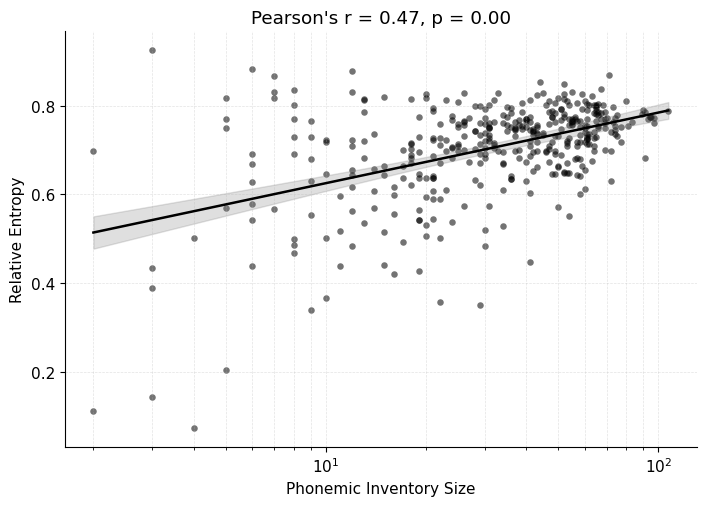}
\end{tabular}
  \caption {{\bf (a)} Rank-frequency plots (note the log-log scale) for the final phoneme frequency distributions in Simulation~1. Each line plots one language. {\bf (b)} Relationship between PIS and relative entropy for the final distributions in Simulation~1.  Each point plots one simulated language. The solid line plots a log-linear regression, and the shading plots its 95\% C.I..}
   \label{fig:sim1}
\end{figure*}

\subsection{Simulation 1: a na\"{\i}ve model}

As a baseline, we simulated the basic model in Section~\ref{sec:formal}, using its simplest instantiation. We considered constant equal probabilities $P_\tau(p)=P_\tau(s)=P_\tau(m)=1/3$ for all timepoints $\tau$. At each time step, the phonemes involved in the change ($\pi_i$ and $\pi_j$ in the notation above) were randomly sampled from the current phoneme inventory with uniform probabilities (i.e., any phoneme was equally likely to be chosen), and the proportion of phonemes involved in the change ($\alpha_\tau$), when necessary, was sampled from a uniform [0,1] distribution. We simulated the evolution of 400 distinct languages, each for 1,000 time steps. Initially, all languages were set to uniform distributions of 34 phonemes (this number was taken as the approximate mean PIS for the world's languages according to PHOIBLE 2.0; \citealp{phoible2}).

As shown in Figure~\ref{fig:sim1}a, even this na\"{\i}ve simulation already generates rank-frequency patterns resembling those observed in empirical phoneme distributions (Figure~\ref{fig:reallang}a). This is so even when all distributions started out as uniforms. However, examining the simulated distributions in more detail reveals crucial differences with real phoneme frequency distributions. Figure~\ref{fig:sim1}b plots the correlation between (log) PIS and the relative entropy of the distribution, which is significant (Pearson's $r=.47,\,p<.01$). However, this goes in exactly the opposite direction than it did in real phoneme distributions: Whereas simulated data show a positive correlation, in actual distributions it is negative. This difference indicates that, in order to capture the properties of phoneme frequencies, our diachronic models need to consider further details.

\begin{figure*}[th]
\begin{tabular}{cc}
{\bf (a)} & {\bf (b)}  \\
  \includegraphics[width=.48\textwidth]{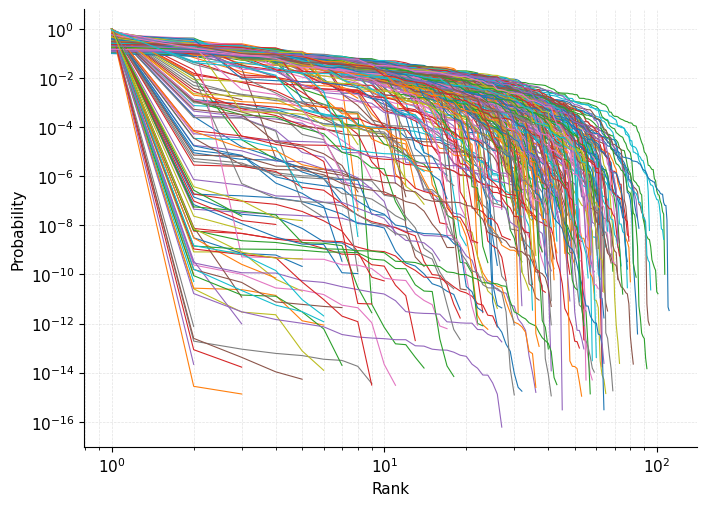} &
  \includegraphics[width=.48\textwidth]{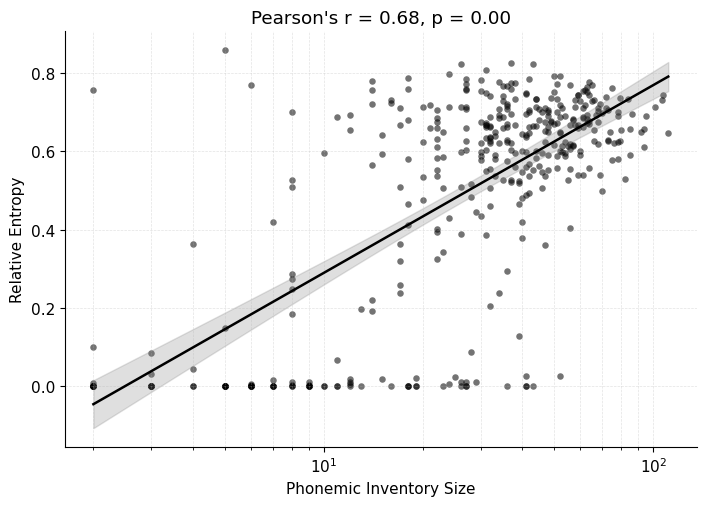} 
\end{tabular}
  \caption {{\bf (a)} Rank-frequency plots (note the log-log scale) for the final phoneme frequency distributions in Simulation~2. Each line plots one language. {\bf (b)} Relationship between PIS and relative entropy for the final distributions in Simulation~2.  Each point plots one simulated language. The solid line plots a log-linear regression, and the shading plots its 95\% C.I..}
   \label{fig:sim2}
\end{figure*}

\subsection{Simulation 2: considering functional load}

Simulation~1 was too na\"{\i}ve in assuming that all phonemes are equally likely to take part in phonological change. This is in fact known not to be the case. The \emph{Functional Load Hypothesis} --tracing back to \citet{Gillieron:1918}-- proposes that phonological contrasts that distinguish many words (i.e., have high functional load) are less likely to be lost through sound change, whereas contrasts that contribute little to lexical differentiation are more prone to merger (for details see, \citealp{Hockett:1955,Hockett:1967}). This hypothesis has received empirical support \citep{Wedel:etal:2013b,Wedel:etal:2013}. Strictly speaking, our models are too coarse-grained to explicitly consider a direct measure of functional load (see \citealp{Surendran:Niyogi:2003}). We can however, indirectly take it to account to some degree: \citet{Wedel:etal:2013} showed that --in general-- a phoneme's functional load is positively correlated with its frequency of occurrence. We can therefore use a phoneme's frequency as a proxy for its functional load.\footnote{\citet{Wedel:etal:2013} showed that phoneme frequency correlates negatively with likelihood of phoneme loss. However, when the effects of actual functional load are disentangled from those of mere frequency, they find that frequent phonemes are \emph{more} likely to be lost. In our simulations, we are making use of the negative correlation that conflates functional load with frequency because of their natural correlation.}

In order to take the functional load hypothesis into consideration, we ran new simulations. As before, we considered constant equal probabilities $P_\tau(p)=P_\tau(s)=P_\tau(m)=1/3$ for all timepoints $\tau$, and the proportion of phonemes involved in the change ($\alpha_\tau$), when necessary, was sampled from a uniform [0,1] distribution. Again, we simulated the evolution of 400 distinct languages, each for 1,000 time steps. Initially, all languages were set to uniform distributions of 34 phonemes.

This time, however, at each time step, the phonemes involved in the change were randomly sampled from the current phoneme inventory with different probabilities. The phonemes that would be split or merged into another one (i.e., the phonemes whose probabilities are decreased in Equations~\ref{eq:primary}, \ref{eq:secondary}, and \ref{eq:merge}) were sampled with probabilities directly proportional to their surprisals (i.e., less frequent phonemes were sampled more often). On the other hand, the phonemes whose probabilities would increase (in Equations~\ref{eq:primary} and \ref{eq:merge}) were sampled uniformly (i.e., all phonemes are equally likely to be chosen). In this way, we have increased the probability that the less frequent phonemes are lost in historical change.

The resulting distributions retain the characteristic shape of phoneme rank-frequency curves observed in real languages, albeit with a substantially increased variance; the introduction of the functional load bias causes a rich-get-richer effect, resulting in many extremely skewed distributions with close to zero relative entropies (see Figure~\ref{fig:sim2}a). The modification did not fix the incorrect sign of the PIS--relative entropy correlation (Pearson's $r=.68,\,p<.01$; see Figure~\ref{fig:sim2}b).

\begin{figure*}[t]
\begin{tabular}{ccc}
  \includegraphics[width=.32\textwidth]{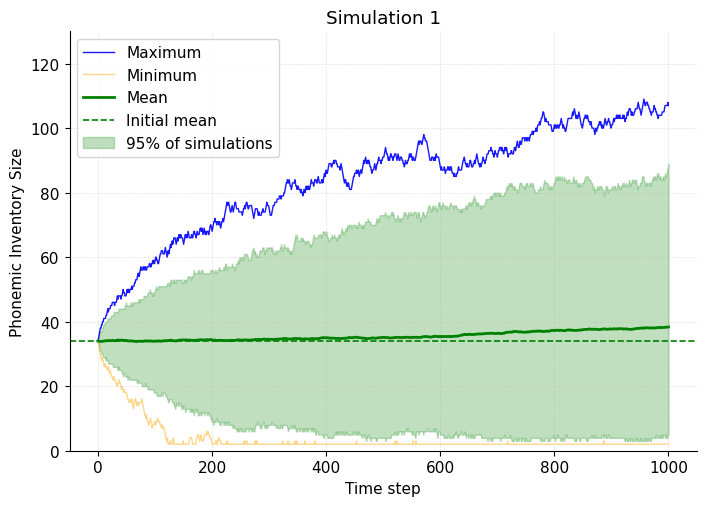} &
  \includegraphics[width=.32\textwidth]{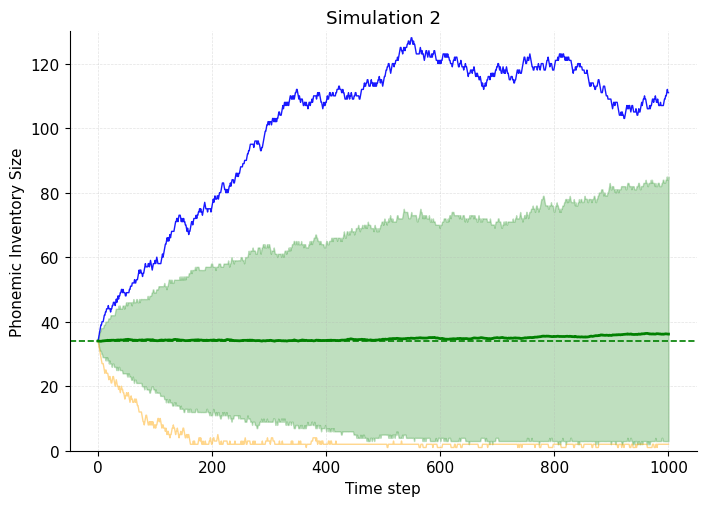} &
  \includegraphics[width=.32\textwidth]{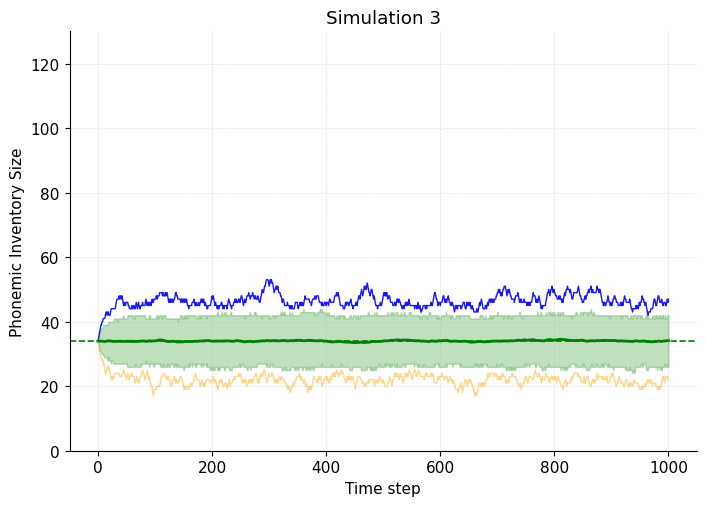}
  \end{tabular}
  \caption{Convergence of the PIS in the simulations. The blue and orange lines respectively plot the maximum and minimum values at each time point across the 400 simulated languages. The solid green lines plot the mean PIS at each time point, the dashed green lines are the predicted mean $\mu$, and the shaded areas contain 95\% of the runs in each of the simulations.}
  \label{fig:sim3evol}
\end{figure*}

\subsection{Simulation 3: adding a central tendency}

Simulation~2 still presents two problems with respect to actual phoneme frequency distributions. First, it does not show the correct relationship between PIS and relative entropy. By itself, this might not be so much of a problem. It could very well be that this correlation truly arises as a consequence of actual compensation. Indeed, \citet{Moscoso:Salhan:2026} find that such correlation could be explained microscopically by compensation at the level of phonotactics, by which languages with larger inventories tend to have phonemes that appear in more predictable contexts. As our models do not include any microscopic details on phonotactics, it could be the case that we are not able to model the emergence of compensation. 

Second, and more critically, in Simulations~1 and 2, PIS values evolve as random walks on the positive integers. At each step, the step size probabilities depend on the probabilities of the three change types. In the idealised case in which $0<\alpha_\tau<1$ almost surely,
\begin{equation} \label{eq:steps}
\begin{aligned}
V_{\tau+1}-V_\tau & = \Delta V_\tau , \\
P(\Delta V_\tau=0)  &= P_\tau(p), \\
P(\Delta V_\tau=+1) &= P_\tau(s), \\
P(\Delta V_\tau=-1) &= P_\tau(m).
\end{aligned}
\end{equation}
If $\alpha_\tau$ can take the boundary value $1$, primary splits and secondary splits may also reduce or preserve $V_\tau$ through elimination of the source phoneme. Moreover, mergers are only defined when $V_\tau$ exceeds a minimal value, which induces a reflecting constraint at small inventories. Whether this random walk has a stationary mean or it has a drift depends on the specific values of the different parameters. However, as long as the probabilities $P_\tau(p)$, $P_\tau(s)$, and $P_\tau(m)$ remain constant in time, the \emph{variance} of $V_\tau$ is bound to increase with time. As a consequence, in this situation, as time goes on, there is no limit to the size that $V_\tau$ can take, and the minimum goes all the way down to two phonemes.\footnote{This was set as a hard limit in the simulations, as it wouldn't make any sense to have languages with a single contrastive phoneme.} The resulting random walks for Simulations 1 and 2 are summarised in Figure~\ref{fig:sim3evol}. Notice that in both simulations, even after 1,000 time steps the ranges of PIS values keep increasing without any upper bounds.

This property is not desirable for a model of phonological change. First, there is no evidence that the range of values in the PIS is increasing with time in the world's languages. Previous work has noted that the number of phonemic contrasts across languages is concentrated within a relatively narrow range. Extremely small or large inventories are rather rare \citep{Maddieson:1984,Anderson:etal:2023}. For instance, according to Phoible v2.0 \citep{phoible2}, the minimum value of the PIS is that of some variants of Rotokas, which may have as few as eleven contrasts (if vowel length is not considered), and the maximum is capped at the 160 contrasts attributed to East Taa. Were the range of PIS left to increase freely in an unconstrained random walk, we would expect more extreme values to be observed among thousands of languages. Also as a result, as can be appreciated comparing Figure~\ref{fig:reallang}a with Figures~\ref{fig:sim1}a and \ref{fig:sim2}a, the variability in the lower rank phonemes is substantially larger in the simulations than in the real language data.

\begin{figure*}[t]
\begin{tabular}{cc}
{\bf (a)} & {\bf (b)} \\
  \includegraphics[width=.48\textwidth]{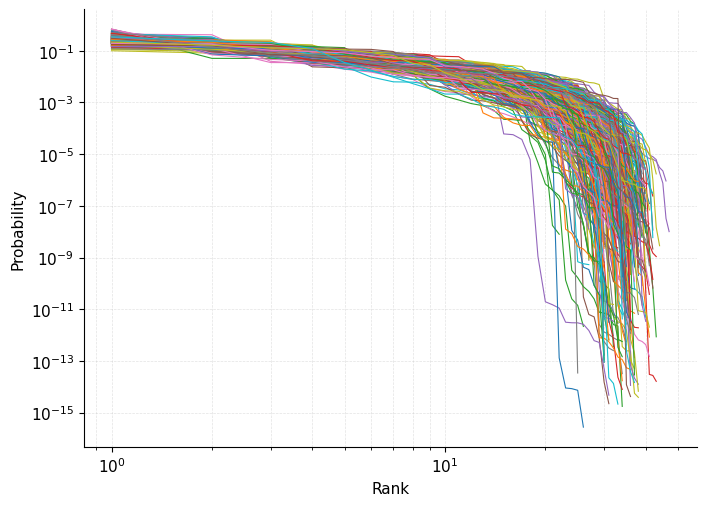} &
  \includegraphics[width=.48\textwidth]{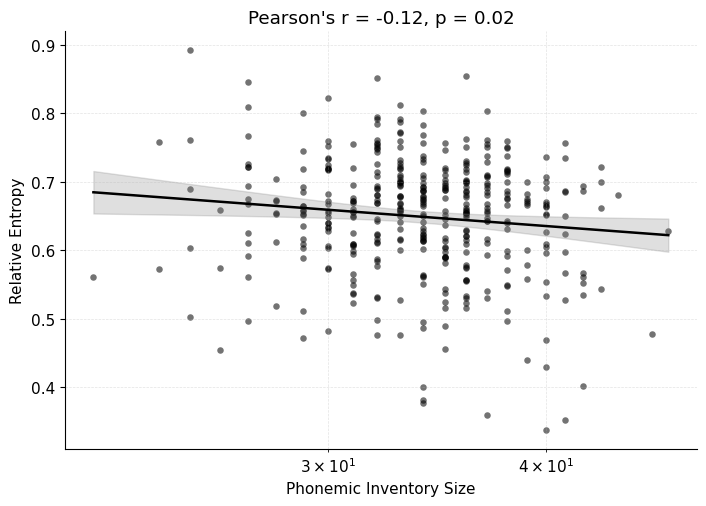} 
\end{tabular}
  \caption {{\bf (a)} Rank-frequency plots (note the log-log scale) for the final phoneme frequency distributions in Simulation~3. Each line plots one language. {\bf (b)} Relationship between PIS and relative entropy for the final distributions in Simulation~3.  Each point plots one simulated language. The solid line plots a log-linear regression, and the shading plots its 95\% C.I..}
   \label{fig:sim3}
\end{figure*}

The only way of avoiding such growing variance of the $V_\tau$ is to make the $P_\tau$ values dependent on the value of $V_\tau$ itself. We assume there is some optimal value of $V_\tau$, which we denote by $\mu$. The further $V_\tau$ is from $\mu$, the less likely it should be to get even further. From Equation~\ref{eq:steps} we see that $P_\tau(s)$ should decrease in value when $V_\tau > \mu$, and in turn, whenever $V_\tau < \mu$, it is $P_\tau(m)$ that should be lowered. We implemented this adaptive strategy using exponential functions, implementing a smooth bias toward $\mu$ while preserving positive probabilities,
\begin{equation}
\begin{aligned}
P_\tau(p) & = \frac{1}{k(\tau)}\\
P_\tau(s) & = \frac{1}{k(\tau)} \, \text{e}^{(\mu-V_\tau)/\mu}\\
P_\tau(m) & = \frac{1}{k(\tau)} \, \text{e}^{(V_\tau-\mu)/\mu} \\
k(\tau) & = 1 + \text{e}^{(\mu-V_\tau)/\mu} + \text{e}^{(V_\tau-\mu)/\mu}
\end{aligned}
\end{equation}
We modified Simulation~2 to include the adaptive scheme above, with $\mu=34$ (the mean PIS across the world's languages). As before, we simulated 400 languages for 1,000 time steps. Figure~\ref{fig:sim3evol} shows that the simulations have now converged in terms of maximum, minimum and range of values of the PIS.

Figure~\ref{fig:sim3}a plots the rank-frequency distributions of the simulated inventories. Notice that the variability on the lower ranks has considerably decreased in relation to Simulations 1 and 2 (compare with Figures~\ref{fig:sim1}a and ~\ref{fig:sim2}a), the distributions are more similar to those observed in real languages (compare with Figure~\ref{fig:reallang}a). Most crucially, as Figure~\ref{fig:sim3}b plots, a significant negative correlation between PIS and relative entropy has now emerged (Pearson's $r=-.12,\,p=.02$). In short, both of the problems in the previous simulations have been solved by introducing the central tendency.

One could fit the different simulation parameters to make the resulting values match human languages more closely, but this is not necessary for the general patterns to arise. Remarkably, the negative relationship between PIS and relative entropy can arise without any actual compensation mechanisms being involved. In our simulations, it is just an unexpected side effect of the interaction between phonological changes, and a central tendency towards a specific number of contrasts. In this sense, what appears as compensation can in fact be epiphenomenal.

\section{Discussion}

The simulations presented here show that several macroscopic properties of phoneme frequency distributions can arise from relatively simple diachronic mechanisms. In particular, stochastic phonological change acting on phoneme inventories naturally generates rank-frequency distributions with exponential tails similar to those observed in real languages. Such distributions arise naturally from the historical processes that continually reshape phonological systems. Our results should be interpreted as providing a generative explanation for the statistical patterns. The goal of the models is not to capture all mechanisms involved in phonological change. Rather, we test whether simple diachronic processes are sufficient to generate the observed macroscopic regularities of phoneme frequency distributions.

Naïve versions of the model fail to capture important empirical patterns. In particular, they predict a positive relationship between phonemic inventory size (PIS) and relative entropy --contrary to the findings of \citet{Moscoso:Salhan:2026}-- and they produce unbounded variation in inventory sizes over time. Introducing a central tendency in the evolution of PIS substantially improves the models. This constrains the random walk behaviour of inventory size, producing stationary dynamics that are more consistent with the observed range of phoneme inventories across languages.

Not including a central attractor --as in Simulations 1 and 2, or in the models of \citet{Ceolin:etal:2019} and \citet{Ceolin:2020}-- results in phonemic inventory sizes that are too broadly distributed. Instead, our assumption of a central tendency in phonemic inventory size is consistent with previous research on phonological systems. Typological surveys have long noted that the number of phonemic contrasts across languages is concentrated within a relatively narrow range, with extremely small or extremely large inventories being rare \citep{Maddieson:1984,Anderson:etal:2023}. At a theoretical level, models such as Adaptive Dispersion Theory propose that phonological systems evolve under competing pressures for perceptual distinctiveness and articulatory economy, which tend to produce stable configurations of contrasts \citep{Liljencrants:Lindblom:1972,Lindblom:1986,Lindblom:Maddieson:1988}. From this perspective, a stabilising tendency in the evolution of phonemic inventories can be interpreted as a coarse-grained reflection of underlying pressures shaping phonological systems. Our model does not attempt to represent these mechanisms explicitly, but instead captures their aggregate effect through a simple probabilistic bias toward a preferred inventory size.

\section{Conclusion}

Taken together, these results suggest that some statistical properties of phonological systems may emerge from the interaction of simple diachronic pressures --in line with the results of \citet{Ceolin:etal:2019} and \citet{Ceolin:2020}-- rather than from direct functional optimization. In particular, the negative relationship between PIS and relative entropy --previously interpreted as evidence for compensatory organisation within the phonological system \citep{Moscoso:Salhan:2026}-- can arise as a by-product of stochastic sound change operating under a stabilising tendency toward a preferred inventory size. This does not rule out the possibility that genuine compensatory mechanisms exist. However, it highlights the importance of considering whether apparent optimisation effects may instead reflect epiphenomenal consequences of the dynamics of sound change.



\bibliography{evol_acl}

\end{document}